\documentclass[10pt,twocolumn,letterpaper]{article}

\usepackage{wacv}
\usepackage{times}
\usepackage{epsfig}
\usepackage{graphicx}
\usepackage{amsmath}
\usepackage{amssymb}
\usepackage{cite}
\usepackage{float}
\usepackage{multirow}
\usepackage{subcaption}
\usepackage{hhline}
\usepackage{makecell}
\usepackage{tabularx}
\usepackage{array}

\usepackage[accsupp]{axessibility}  
\usepackage[pagebackref=true,breaklinks=true,colorlinks,bookmarks=false]{hyperref}

\newcommand{\tb}{\textbf}
\newcommand{\ti}{\textit}

\makeatletter
\def\thickhline{%
  \noalign{\ifnum0=`}\fi\hrule \@height \thickarrayrulewidth \futurelet
   \reserved@a\@xthickhline}
\def\@xthickhline{\ifx\reserved@a\thickhline
               \vskip\doublerulesep
               \vskip-\thickarrayrulewidth
             \fi
      \ifnum0=`{\fi}}
\makeatother

\newlength{\thickarrayrulewidth}
\def\wacvPaperID{14} 

\wacvfinalcopy 

\ifwacvfinal
\fi

\ifwacvfinal
\usepackage[breaklinks=true,bookmarks=false]{hyperref}
\else
\usepackage[pagebackref=true,breaklinks=true,colorlinks,bookmarks=false]{hyperref}
\fi

\begin{document}


\title{Hyperspectral Image Super-Resolution in Arbitrary Input-Output Band Settings}

\author{Zhongyang Zhang\footnote[1]{}
\and
Zhiyang Xu\footnote[1]{}
\and
Zia Ahmed\footnote[2]{}
\and
Asif Salekin\footnote[3]{}
\and
Tauhidur Rahman\footnote[1]{} \and 
University of Massachusetts Amherst\footnote[1]{} \and

University at Buffalo\footnote[2]{} \and
Syracuse University\footnote[3]{} \and {\tt\small \{zhongyangzha,zhiyangxu,trahman\}@cs.umass.edu} \and
{\tt\small zahmed2@buffalo.edu} \and {\tt\small asalekin@syr.edu}
}

\maketitle
\thispagestyle{empty}

\begin{abstract}

Hyperspectral image (HSI) with narrow spectral bands can capture rich spectral information, but it sacrifices its spatial resolution in the process. Many machine-learning-based HSI super-resolution (SR) algorithms have been proposed recently. However, one of the fundamental limitations of these approaches is that they are highly dependent on image and camera settings and can only learn to map an input HSI with one specific setting to an output HSI with another. However, different cameras capture images with different spectral response functions and bands numbers due to the diversity of HSI cameras. Consequently, the existing machine-learning-based approaches fail to learn to super-resolve HSIs for a wide variety of input-output band settings. We propose a single Meta-Learning-Based Super-Resolution (MLSR) model, which can take in HSI images at an arbitrary number of input bands' peak wavelengths and generate SR HSIs with an arbitrary number of output bands' peak wavelengths. We leverage NTIRE2020 and ICVL datasets to train and validate the performance of the MLSR model. The results show that the single proposed model can successfully generate super-resolved HSI bands at arbitrary input-output band settings. The results are better or at least comparable to baselines that are separately trained on a specific input-output band setting. 


\end{abstract}

\section{Introduction}

Hyperspectral imaging has proven effective in solving numerous computer vision tasks, including image segmentation, object recognition, material sensing, and surface characterization in different domains like remote sensing, astronomy, materials science, and biology \cite{MFauvelSpectralSpatialClassification2013,CKwanSpectralUnmixingClassification2006,HienVanTrackingObjectReflectance2010,YTarabalkaSegmentationClassificationHyperspectral2010}. HSI can capture rich spectral information by capturing bands near multiple peak wavelengths with a narrow bandwidth. However, HSI is often in low spatial resolution \cite{yokoyaHyperspectralMultispectralData}, as more filters need to be accommodated in the optical sensor mosaic. This drawback hinders the use of HSI for applications that require high resolution (HR) HSI. Many efforts have been devoted to the study of HSI SR. Among all of them, the fusion-based HSI SR recently gains a lot of attention due to its outstanding performance and accessibility of HR RGB. It leverages simultaneously collected HR multispectral image to spatially super-resolve the low resolution (LR) HSI \cite{weiHyperspectralMultispectralImage2014,simoesConvexFormulationHyperspectral2015,kanatsoulisHyperspectralSuperResolutionCoupled2018,heHyperspectralSuperResolutionCoupled2020,wanNonnegativeNonlocalSparse2020,xuNonlocalCoupledTensor2020}. Both optimization based and deep-learning based approaches have been proposed for fusing high spatial frequency information in RGB with LR HSI to generate HR HSI.

\begin{figure}[t]
\begin{center}
\includegraphics[width=\linewidth]{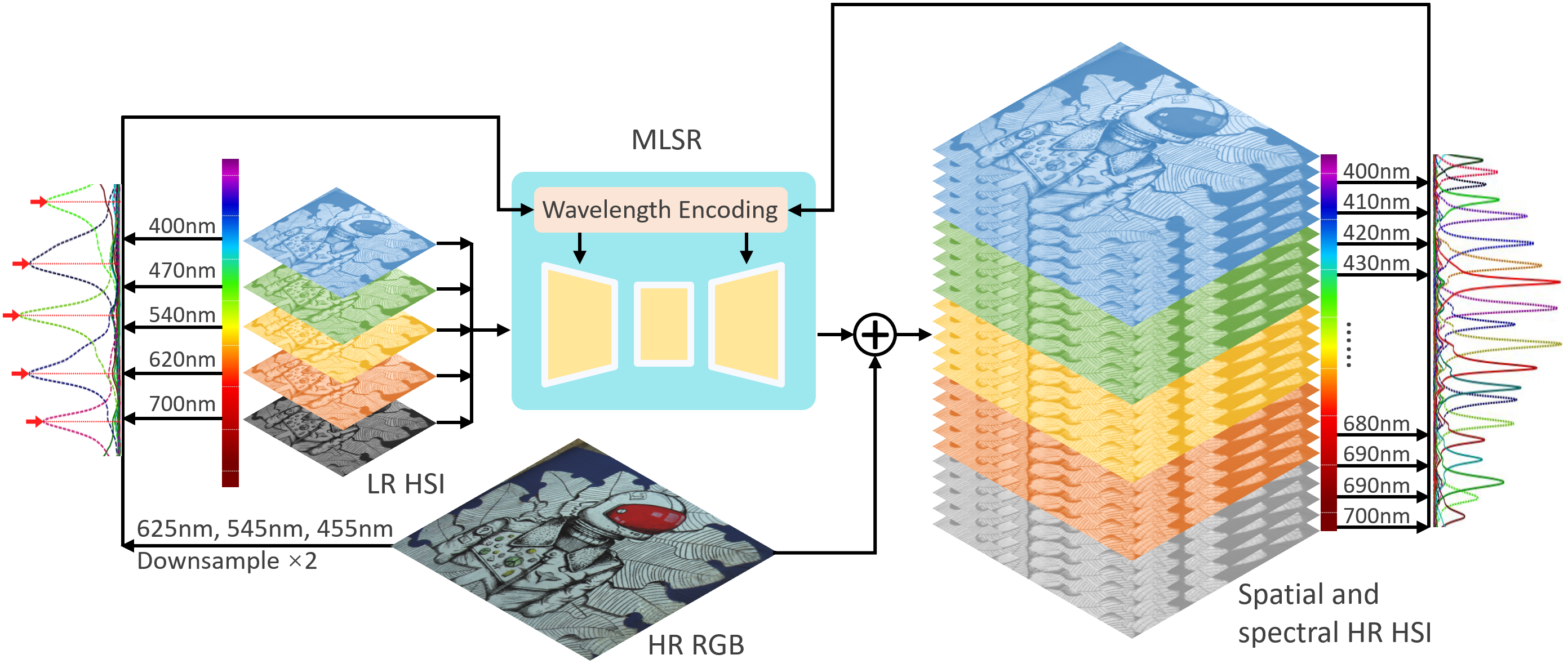}
\end{center}
\vspace{-0.3cm}
\caption{The overall architecture of our proposed MLSR. It takes in an HSI with one bands setting, super-resolve it with RGB guidance, and generate HR HSI with another bands settings. The Red arrows on the left spectral response curves represents the peak wavelengths.}
\label{fig:cover}
\end{figure}

Despite the moderate success of fusion-based HSI SR, it is limited to fixed input-output band settings, making it impractical in real-world settings. Different HSI cameras can image at different peak wavelengths. Hence, existing SR models that are trained to operate with a fixed set of input-output wavelengths are often fairly restrictive for broad application with other HSI cameras \cite{ zhangDeepBlindHyperspectral2020a, weiDeepRecursiveNetwork2020a, changHSIDeNetHyperspectralImage2019}. This project proposes a meta-learning-based SR model to address those two limitations. Meta-learning\cite{hospedalesMetaLearningNeuralNetworks2020} is a broad definition of a wide range of methods that can learn to learn. In this paper, instead of MAML-like methods, "meta" refers to a module in the proposed structure whose weight is the output of another network that takes in other information as input. The question we want to answer in this paper is: ``Can a single meta-learning-based machine learning model perform SR on HSI with bands at arbitrary input peak wavelengths and output SR image bands at another arbitrary wavelengths setting.'' This work answers this question by demonstrating how a single SR model can be trained on HSIs with different input-output band numbers and peak wavelengths. The core component of MLSR is a meta weight prediction module that maps from wavelengths to the weights of the meta wavelengths convolution layer. The meta block takes the responsibility of fusing and unmixing spectrum information during the encoding and decoding of different input-output bands, which frees the backbone from taking care of both spatial SR and spectrum encoding/decoding.

This paper introduces a new task: RGB-guided HSI super-resolution with arbitrary input-output bands. This task requires flexible input formats, spectral-domain understanding, retrieval at arbitrary spectrum position, and super-resolved reconstruction. We systematically train and evaluate our proposed MLSR on multiple datasets (i.e., NTIRE 2020 and ICVL) and compare it with the state-of-the-art spatial and spectral joint super-resolution approaches \cite{meiSpatialSpectralJoint2020}. The main highlight of our approach is that only one single trained model can achieve SR at arbitrary input-output HSI bands settings while the standard deep-learning methods need to train one model for each setting. In most cases, the proposed MLSR outperforms baseline methods.



\section{Related Works}

\subsection{HSI Super-resolution}

Existing HSI SR algorithms can be broadly classified into two approaches: single HSI SR and MSI-HSI fusion-based SR. The former assumes that no auxiliary MSI image like HR RGB is available and solely operate on the input LR HSI data \cite{akgunSuperresolutionReconstructionHyperspectral2005, heSuperresolutionReconstructionHyperspectral2016, meiHyperspectralImageSpatial2017, liSingleHyperspectralImage2018, xieHyperspectralImageSuperResolution2019, arunCNNBasedSuperResolutionHyperspectral2020, huHyperspectralImageSuperResolution2020, wangSpatialSpectralResidualNetwork2020}.
Unlike single-image scheme, fusion-based methods take advantage of both the high spatial resolution in the multispectral image (MSI), and the high spectral resolution in the HSI. The core idea is to guide the HSI SR with the high spatial resolution information captured by the MSI images. Fusion approaches include Bayesian-based approaches \cite{akhtarBayesianSparseRepresentation2015a, loncanHyperspectralPansharpeningReview2015, changWeightedLowrankTensor2017, bungertBlindImageFusion2018}, 
tensor-based approaches \cite{dianHyperspectralImageSuperResolution2017, zhangSpatialSpectralGraphRegularizedLowRank2018, xuNonlocalPatchTensor2019, heHyperspectralSuperResolutionCoupled2020, wanNonnegativeNonlocalSparse2020}, 
matrix factorization-based approaches \cite{weiMultiBandImageFusion2016, dianHyperspectralImageSuperResolution2019, borsoiSuperResolutionHyperspectralMultispectral2020a, liuTruncatedMatrixDecomposition2020}, and deep-learning-based methods\cite{ zhangDeepBlindHyperspectral2020a, weiDeepRecursiveNetwork2020a, changHSIDeNetHyperspectralImage2019, zhangHyperspectralImageReconstruction2019, jiangLearningSpatialSpectralPrior2020a}. Moreover, researchers have been studying the application of image prior in this domain \cite{sidorovDeepHyperspectralPrior2019, jiangLearningSpatialSpectralPrior2020a}.


\subsection{Spectral Interpolation}

Spectral interpolation has been explored extensively in spectroscopy for estimating the reflectance or transmittance spectra where fine-grained measurement is not available.
To this end, several interpolation methods including linear, cubic, PCA-based methods have proven effective for spectral interpolation \cite{huynhComparativeEvaluationSpectral2013, westlandInterpolationSpectralData2014}. Furthermore, according to \cite{sandorSpectralInterpolationErrors2005}, cubic interpolation has been established as an effective and adequate spectral interpolation technique, which shows that a more nonlinear model for interpolation does not boost the performance any further. More recently, deep-learning-based methods have also been proposed for jointly learning spectral-spatial super-resolution \cite{meiSpatialSpectralJoint2020}. Recent works have also explored RGB to HSI mapping with deep-learning and achieved remarkable performance \cite{aradNTIRE2020Challenge2020, shoeibyPIRM2018ChallengeSpectral2019, lahoudMultimodalSpectralImage2019}. However, they can only work on RGB to produce HSI with specific settings and do not have any input-output flexibility.

\subsection{Meta Strategy in Image Super-Resolution}

Meta-learning\cite{hospedalesMetaLearningNeuralNetworks2020} has attracted significant attention in recent years. Weight prediction network or hypernetworks is an essential component of meta-learning\cite{hospedalesMetaLearningNeuralNetworks2020}. Hypernetworks\cite{haHyperNetworks2016, brockSMASHOneShotModel2017} are neural networks that predict the weight of another neural network based on some auxiliary information or embeddings. Hypernetworks are usually used in multi-task learning and synthesizing predictive models by conditioning an embedding of the support dataset \cite{rakellyEfficientOffPolicyMetaReinforcement2019a, rusuMetaLearningLatentEmbedding2019}. In this scenario, the target network's weights are not learned during the training process; instead, it is predicted by another network. For example, Hu \etal \cite{huMetaSRMagnificationArbitraryNetwork2019} proposed a meta-upsample module for SR tasks and made the upsample scale arbitrary and continuous. Cai \etal\cite{caiMemoryMatchingNetworks2018} used a hypernetwork to predict the classifier's weight to migrate to new categories without re-train few-shot learning. Our proposed MLSR also applied this mechanism which will be explained in the next section.

\begin{figure*}[t!]
\begin{center}
    \includegraphics[width=0.95\linewidth]{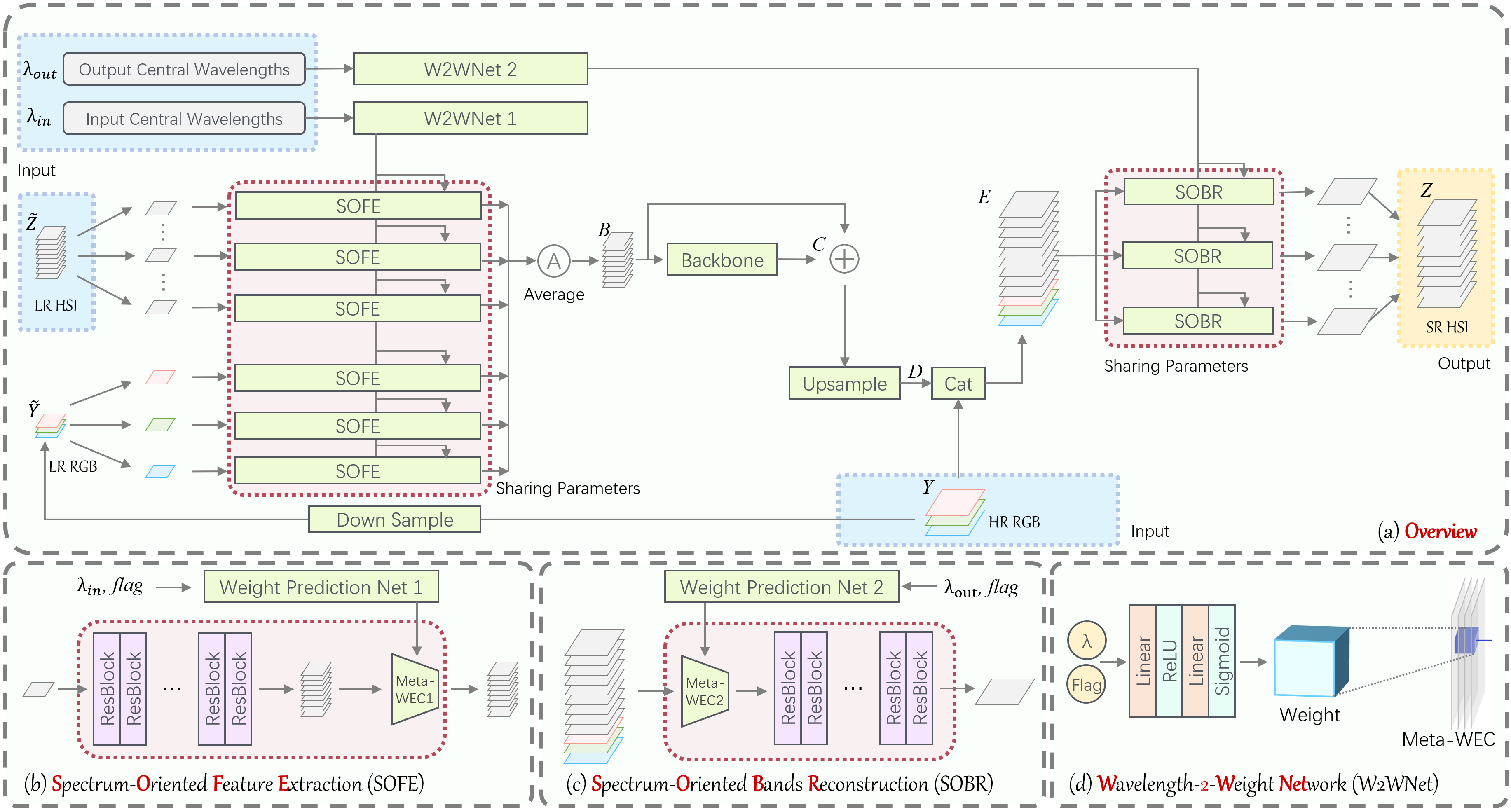}
\end{center}
\vspace{-0.5cm}
\caption{The architecture of MLSR. (a) The overview of the whole model. (b) Structure of SOFE. (c) Structure of SOBR. (d) Structure of W2WNet. MLSR consists of three stages: SOFE in the head, backbone and upsample in the body and SOBR in the tail. Light blue dashed rectangles surround inputs, and the golden dashed rectangle marks the output.}
\label{fig:overview}
\end{figure*}

\subsection{End-to-End SSSR}

There are several works\cite{meiSpatialSpectralJoint2020, yiJointSpatialspectralResolution2020} exploring spatial and spectral SR recently. They find that compared to spatial and spectral SR separately, the end-to-end model can avoid spectral distortion and spatial inconsistency and lead to better overall results. Yi's method \cite{yiJointSpatialspectralResolution2020} bases on optimization, while Mei \etal \cite{meiSpatialSpectralJoint2020} focuses on deep-learning models and discusses about the best way to combine spatial and spectral SR stages. As they show us some good results, these methods are fixed to a certain strict setting and cannot be applied on multiple datasets with different settings.

\section{MLSR}

In this section, we introduce the proposed MLSR. Fig. \ref{fig:overview} shows the overall architecture. It mainly consists of five parts: Spectrum-Oriented Feature Extraction (SOFE), Wavelength-2-Weight Network (W2WNet), Spectrum-Oriented Bands Reconstruction (SOBR), a backbone network, and an upsampler.

\subsection{Formulation and Motivation}\label{MOT}

The overall architecture is motivated by performing SR on HSIs with different input settings and output HR bands on demand. A feature extraction layer with fixed parameters does not have the flexibility to handle input with variable channels. To precisely extract features in each band, we need a meta module with a full understanding of wavelengths information to predict the feature extractor corresponding to each image band. Each band has a unique spectral response curve. Since the peak wavelengths are the most prominent properties for each band, we use it as a identifier in this work for simplicity. Our design also takes advantage of fusion-based HSI SR by incorporating high-resolution spatial information from HR RGB which guides LR HSI during the SR process. The backbone is formed by residual dense blocks \cite{zhangResidualDenseNetwork2018} which shows superior performance on RGB SR.
Like the feature extraction layer, the output layer also leverages meta-learning to effectively reconstruct an arbitrary number of image bands given the corresponding peak wavelengths.

We denote the low spatial and spectral resolution image as $\tilde{Z} \in \mathbb{R}^{h\times w \times s}$ and the LR RGB image as $\tilde{Y} \in \mathbb{R}^{h\times w \times 3}$ where $h$,$w$, and $s$ stand for height, width and number of LR HSI bands. We denote the high spatial and spectral resolution HSI as $Z \in \mathbb{R}^{H\times W \times S}$ and the HR RGB image as $Y \in \mathbb{R}^{H\times W \times 3}$, where $H$,$W$, and $S$ stand for height, width and number of HR HSI bands. Moreover, as peak wavelengths of the input and output spectral bands are made variable in our method, they are denoted as $\lambda_{in} \in \mathbb{R}^{s+3}$, where $\lambda_{in}$ =$[\lambda_{hsi},\lambda_{rgb}], \lambda_{hsi} \in \mathbb{R}^{s}, \lambda_{rgb} \in \mathbb{R}^{3}$ , [,] denotes concatenation operation, and $\lambda_{out} \in \mathbb{R}^{S}$. The $\lambda_{hsi}$ and $\lambda_{rgb}$ are input wavelengths of the LR HSI and RGB bands.

The objective of MLSR is solving:
\begin{align}
Z = g(\tilde{Z}, Y, \lambda_{in}, \lambda_{out}|\phi, \theta)
\end{align}
where $g$ stands for the overall model inference, $\phi$ represents the weight of the hypernetwork W2WNet, and $\theta$ represents parameters of all other parts in MLSR.

\subsection{W2WNet}\label{W2WNet}

Given peak wavelengths and bandwidth indicators, W2WNet is to predict the weights of the meta wavelength embedding convolution (Meta-WEC) module. During training, the spectrum information naturally embedded in each image band can be encoded into the parameters of the W2WNet and extracted at inference. This design allows flexibility in the input/output formats. It enables the Meta-WEC to filter out noise from other bands and keep distinct features for a particular band in extraction and reconstruction. The meta module in W2WNet proves the possibility of building a general cross-dataset model for HSI images. 

The input of W2WNet is two scalars: corresponding peak wavelength of an HSI band and a binary digit indicating whether this band is from an HSI or RGB image. In our experiment, a 1x1 convolution layer in Meta-WEC can filter out noisy information and extract bands' features. To make the model compact and highly efficient, we decide to use the 1x1 convolution layer. Better performance can be achieved by swapping in a larger module. We use this simple network to demonstrate the advances of our meta-network. Since the required storage of information is small, we used two fully-connect layers to store the bands' knowledge. Fig. \ref{fig:overview}(d) shows the structure of the W2WNet, where the ``weight" block represents the predicted weights for the Meta-WEC. W2WNet1 and W2WNet2 take input and output wavelengths, respectively.

\subsection{SOFE}\label{SOFE}
SOFE generates feature representation from each input band corresponding to its specific spectrum information (i.e., wavelength and bandwidth). It consists of a feature extraction module and a meta wavelength embedding convolution (Meta-WEC) module. Meta-WEC follows a similar Meta mechanism as \cite{huMetaSRMagnificationArbitraryNetwork2019}. The weights of Meta-WEC are the output of W2WNet.

We loop through the input LR HSI bands and apply our feature extractor to each band independently. The feature extractor module consists of eight Residual Blocks. After extracting the generic features, we use the Meta-WEC module, a 2D convolution with a $1 \times 1$ kernel, to encode wavelengths information into the feature of each band correspondingly. This operation essentially embeds a band's spectrum information into its feature representation. 
This design makes the model more robust in terms of input-output formats.

Additionally, as the bands in RGB image has wider bandwidth and embeds spectral information in a wider range than narrow-band HSI bands, following the previous work \cite{huHyperspectralImageSuperResolution2020}, we pass each band of the paired LR RGB to SOFE in parallel as well. After SOFE, we merge the spectrum-informed feature maps for each band by taking the average to form a general embedding. 
When we train our model on five input bands, LSTM learns that band 0 is followed by bands 7. However, when we train our model on seven input bands, band 0 is followed by band 5. The LSTM is not able to correctly capture the correlation between image bands.


We can calculate the spectrum-informed feature maps generated by SOFE modules as follow:
\begin{align}
    B &= \frac{1}{s+3}\sum_{i=1}^{s+3}Conv(R_0([\tilde{Z},\tilde{Y}]_i),W_{in}^i)
\end{align}
where $Conv$ denotes Meta-WEC1, $B \in \mathbb{R}^{h \times w \times G}$ stands for the spectrum-informed feature map, $W_{in}^i$ represents the weights of Meta-WEC for each LR HSI or LR RGB band where $i$ is the index of the concatenated LR HSI and LR RGB bands, $R_0$ means the feature extraction module in SOFE, and G represents the channel number of the feature map extracted in the last step. $Conv$ denotes the Meta-WEC applied on the features of $i^{th}$ band and takes matrix $W_{in}^i$ as its weights. $W_{in}^i$ is the output from W2WNet1 with wavelength $\lambda_{in}^i$ as its input.

\subsection{Backbone and Upsampler}\label{BB}

The backbone uses a deep neural network to learn deep features and map LR representation $B$ to HR representation $D$ (figure \ref{fig:overview}). We use eight Residual Dense Blocks (RDB) proposed by Zhang \etal\cite{zhangResidualDenseNetwork2018} as our backbone module. This module is widely adopted in all kinds of SR tasks and has shown strong performance. We add a skip connection over the whole backbone to resolve the gradient vanishing problem caused by the deep network\cite{heDeepResidualLearning2015,zhangResidualDenseNetwork2018}. We use eight layers of RDB in the backbone, which is half of the layers than the original proposed Residual Dense Network.


We use sub-pixel upsampling layer proposed by Shi \etal in \cite{shiRealTimeSingleImage2016} as our upsampler. After upsampling, the general embedding $C$ becomes $D \in \mathbb{R}^{H \times W \times G}$, which have the same spacial size as $Z$. Then $D$ and $Y$ are concatenated together along the channel dimension, and the result feature map $E$ contains both high spectral resolution information from $D$ and high spatial information from $Y$.

The feature map $D$ is a general embedding which contains the HR bands' knowledge for all covered spectrum wavelength range, and the module SOBR can be viewed as a filter to extract a single band at a given wavelength. 

After the upsampling module, the feature map $D$ already has the same spatial size as HR RGB. As HR RGB contains more fine-grained HR information and can provide guidance for the final output, feature map $D$ is concatenated with it before going to SOBR.

\subsection{SOBR}\label{SOBR}

SOBR reconstructs the output image bands guided by required spectrum (i.e., wavelength and band) information, backbone generated HR embedding, and high-resolution spatial information from HR RGB. SOBR consists of Meta-WEC2 and a reconstruction module consisting of four Residual Blocks.

SOBR extract features from the upsampled general embedding for each output band given its peak wavelength. W2WNet generates the weights for Meta-WEC2 based on the wavelengths. The output of SOBR is a single band of HR HSI, and SOBR generates $S$ bands for the SR HSI. The operation in SOBR can be represented as follow:
\begin{align}
    Z &= Cat_{k=1}^SR_1(Conv([D, Y],W_{out}^k))
\end{align}
where $Cat$ denotes the concatenate operation, $Conv$ denotes Meta-WEC2. $D$ denotes the upsampled output of the backbone module, $R_1$ stands for the reconstruction module in SOFE, and $W_{out}^k$ is the weight of Meta-WEC2 for $k^{th}$ output band. We apply the reconstruction model after Meta-WEC2 to alleviate the high-level discrepancies in the reconstructed features.

\section{Experiments}

\subsection{Datasets}\label{datasets}

Most of the popular HSI datasets are poorly organized and lack diversity due to their limited size and object types. Because of its generalizability and flexibility, the proposed MLSR can not be properly trained on those tiny datasets. Thus mainly two datasets are compared on: NTIRE2020 \cite{aradNTIRE2020Challenge2020} and ICVL \cite{aradSparseRecoveryHyperspectral2016a}, both of which contain more than 200 images. 


\textbf{NTIRE2020:} The NTIRE 2020 dataset \cite{aradNTIRE2020Challenge2020} is currently considered as the most comprehensive HSI dataset. The HSIs are captured by a Specim IQ mobile hyperspectral camera. At every 10nm between 400nm and 700nm, an image band is captured by a distinct filter. Thus an image in NITRE has 31 bands with $512 \times 482$ spatial dimension.

\textbf{ICVL:}
The ICVL dataset was released in 2016 \cite{aradSparseRecoveryHyperspectral2016a}. The hyperspectral images were captured by the Specim PS Kappa DX4 hyperspectral camera. Each original HSI consists of 519 bands from 400nm to 700nm with a step size of 1.25nm. The 519 bands are downsampled to 31 bands, and the step size becomes 10nm. The dimension of processed HSI in the ICVL dataset is $1392 \times 1300 \times 31$.


\subsection{Training Details}

As this paper focuses on HSI SR with arbitrary input-output settings, the experiments are carefully designed to support the claim. There are three types of experiments in total, which are: 1. Use evenly sampled LR input bands to predict all possible SR bands. 2. Use completely randomly sampled input bands to predict all possible SR bands. 3. Use central LR bands to predict SR bands on the sides. Since the first experiment is more standardized and easier to compare with baseline methods, it is selected as the main experiment, while the other two are studied in the auxiliary experiments subsection. The rest of this subsection are details of the main experiment.

MLSR is trained and evaluated on two datasets, NTIRE2020 and ICVL. For NTIRE2020, we follow the same organization of training and evaluation datasets of the NTIRE2020 competition \cite{aradNTIRE2020Challenge2020}. For ICVL, 40 images are selected as evaluation set, while the rest are training set. 

For each dataset, five sub-datasets are made by evenly sampling five to nine bands from all 31 bands in the original LR HSI images. A series of sub-datasets are made for each dataset to train MLSR as they share the same domain knowledge while yet have different input band number and peak wavelengths. 

Since each output band is generated independently, the output bands' peak wavelengths are set to the same as HR HSI, 31 in our case, to let W2WNet2 learn as much output pattern as possible. For one trained model, it can be applied to various input bands' combinations and output HR HSI bands at any requested peak wavelengths. All sub-datasets of a dataset are trained simultaneously and each batch is randomly sampled from one of these sub-datasets. During the evaluation, one sub-dataset is used each time to generate a PSNR/SSIM pair. The SR capability of model is evaluated on multiple scales: $\times 2$, $\times 3$ and $\times 4$.

Each LR input image pair(RGB and HSI) is cropped to a $50 \times 50$ patch randomly while keeping the contents of RGB and HSI the same. For data augmentation, each patch is applied $90^\circ$ rotation, horizontal and vertical flip randomly with $50\%$ possibility. All the models are trained from scratch since the input channel number is not compatible with normal RGB models, and thus pre-train technique cannot be applied. The learning rate is set to $10^{-4}$ initially and will decrease by half every 20 epochs. Moreover, for loss function, we followed previous works\cite{limEnhancedDeepResidual2017, huMetaSRMagnificationArbitraryNetwork2019} and applied L1 loss for better convergence. The graphic cards we use are eight RTX 2080ti or RTX 1080ti with 11GB of video memory. The corresponding batch size for $\times 2$, $\times 3$ and $\times 4$ are 64, 32, and 32. Each model is trained for approximately 200 epochs, and training takes one and a half days, two days, and two and a half days for these three scales.

\begin{table*}[ht!]
\begin{center}
\begin{tabular}{l c c c c c c c c c c c c}
\thickhline
\\[-1em]
& & & \multicolumn{3}{c}{\textbf{NTIRE2020}} &  \multicolumn{3}{c}{\textbf{ICVL}}\\ 
\\[-1em]
\hhline{~~~------}
\makecell{Stage1} & \makecell{Stage2} & \makecell{Input\\Bands} & \makecell{ Scale $\times 2$ \\ PSNR/SSIM} & \makecell{ Scale $\times 3$ \\ PSNR/SSIM} & \makecell{ Scale $\times 4$ \\ PSNR/SSIM} & \makecell{ Scale $\times 2$ \\ PSNR/SSIM} & \makecell{ Scale $\times 3$ \\ PSNR/SSIM} & \makecell{ Scale $\times 4$ \\ PSNR/SSIM}\\
\hline\hline
Linear & EDSR    & \multirow{7}{*}{5} & 37.26/0.973 & 32.92/0.929 & 31.31/0.897 & 44.69/0.991 & 42.80/0.983 & 42.18/0.978\\
Linear & RDN     &  & 38.62/0.981 & 35.76/0.956 & 35.41/0.948 & 44.80/0.992 & 43.06/0.984 & 42.43/0.979\\
Cubic  & EDSR    &  & 35.66/0.960 & 31.72/0.917 & 30.33/0.882 & 43.48/0.990 & 41.66/0.981 & 40.77/0.975\\
Cubic  & RDN     &  & 36.35/0.968 & 34.05/0.941 & 33.94/0.931 & 43.62/0.990 & 41.47/0.981 & 40.71/0.976\\

\multicolumn{2}{c}{SepSSJSR1$^*$} &  & 40.48/0.972 & 31.68/0.848 & 32.29/0.853 & \tb{48.08}/0.992 & 44.28/0.980 & 43.18/0.973\\
\multicolumn{2}{c}{SimSSJSR$^*$} &  & 37.63/0.949 & 30.27/0.796 & 30.23/0.786 & 44.24/0.980 & 39.82/0.955 & 39.27/0.943\\

\multicolumn{2}{c}{\tb{MLSR}} &  & \tb{41.77/0.985} & \tb{39.56/0.966} & \tb{38.07/0.966} & 47.89/\tb{0.993} & \tb{46.59/0.992} & \tb{45.94/0.991}\\
\hline
Linear & EDSR    & \multirow{7}{*}{7} & 38.52/0.976 & 33.38/0.930 & 31.56/0.900 & 45.02/0.993 & 43.04/0.984 & 42.37/0.979\\
Linear & RDN     &  & 39.30/0.984 & 36.89/0.959 & 36.55/0.951 & 45.34/0.993 & 43.33/0.985 & 42.65/0.980\\
Cubic  & EDSR    &  & 39.06/0.975 & 33.47/0.930 & 31.60/0.898 & 45.34/0.993 & 43.21/0.984 & 42.51/0.979\\
Cubic  & RDN     &  & 40.37/0.984 & 36.97/0.959 & 36.70/0.951 & 45.55/0.993 & 43.34/0.985 & 42.72/0.980\\

\multicolumn{2}{c}{SepSSJSR1$^*$} & & 41.23/0.973 & 31.94/0.852 & 32.51/0.856 & \tb{48.85/0.993} & 44.31/0.980 & 43.23/0.972\\
\multicolumn{2}{c}{SimSSJSR$^*$} &  & 39.13/0.960 & 30.79/0.812 & 30.75/0.805 & 44.83/0.982 & 41.04/0.959 & 39.90/0.950\\

\multicolumn{2}{c}{\tb{MLSR}} &  & \tb{42.06/0.985} & \tb{39.57/0.966} & \tb{38.12/0.966} & 48.03/0.993 & \tb{46.58/0.992} & \tb{46.10/0.991}\\
\thickhline
\end{tabular}
\end{center}
\vspace{-0.4cm}
\caption{Evaluation results on NTIRE2020 and ICVL. \ti{Step1} and \ti{Step2} stand for the spectral interpolation and spatial super-resolution method, respectively. \tb{*} denotes methods that train one model for each setting.}
\label{tab:results}
\end{table*}

\subsection{Baseline methods}

To the best of our knowledge, there is no HSI super resolution algorithm that can support arbitrary input-output band settings. To compare the performance of MLSR, we carefully developed several baseline methods that have been trained and tested on specific input-output band settings (i.e., 5 or 7 specific input bands to all 31 bands). Baselines can learn a specific input-output setting better since it does not have to learn to do SR in a new setting.
In Table \ref{tab:results}, baselines can be classified into two categories: two-stage methods and one-stage methods. The former do spectral and spatial SR separately, while the latter do both at the same time.

For deep-learning-based models, SepSSJSR1 and SimSSJSR\cite{meiSpatialSpectralJoint2020} are selected. SepSSJSR1 is a two-stage method, while SimSSJSR is a one-stage method. They are completely retrained on the same datasets using the same training method during the experiments. Since the two deep-learning-based methods have fixed structures and can only be trained and evaluated on a fixed input setting, one separate model needs to be trained for each band setting, although it is not fair for MLSR. 

As similar works are limited even with simplification, besides methods mentioned above, several original baselines are designed as well. Each baseline method's input needs to be flexible to compare impartially, especially the band's number. However, all existing parametric methods, including deep-learning-based ones, cannot provide this flexibility. Therefore, some non-parametric methods have to be selected. They interpolate inputs with various bands number to a fixed higher spectral resolution LR HSI and then send the result to a following spatial SR model to generate the HSI with high spectral and spatial resolution. As a result, those original baselines are all two-stage methods in spectral and spatial SR order.

In the first stage, Cubic and Linear are chosen since they 
can guarantee the ``one pipeline for all settings" target. Each pixel in the input HSI is treated as a spectrum curve, and the value in different bands are some sample points on the curve. The spectral interpolation aims to resample more densely on the curve. In the second stage, a modified version of Enhanced Deep Super-Resolution network (EDSR)\cite{limEnhancedDeepResidual2017}, and Residual Dense Network (RDN)\cite{zhangResidualDenseNetwork2018} are applied. Since they do not use HR RGB, for the fairness of comparison, they are both modified to fusion-based methods similarly to MLSR. LR RGB bands are concatenated with LR HSI initially, and the HR RGB bands are concatenated with the output feature map after the upsampling module in the tail. EDSR and RDN are selected because they have shown superior RGB SR performance, and both are widely used as baselines in recent SR papers. Moreover, as the backbone of MLSR is a modified version of RDN, using it as the second stage baseline model can better support the effectiveness of the proposed  structure.

\begin{figure}[ht]
\begin{center}
\includegraphics[width=\linewidth]{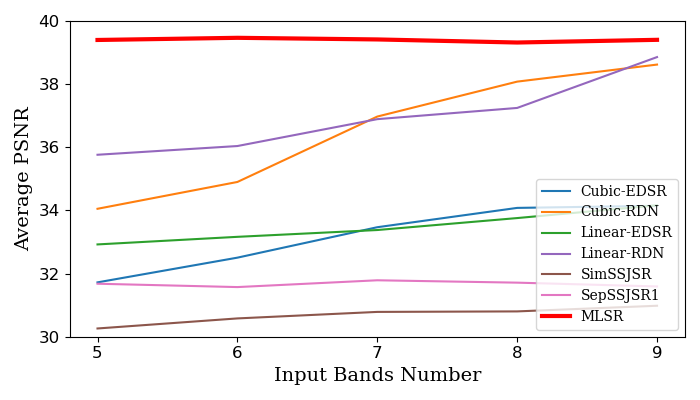}
\end{center}
\vspace{-0.3cm}
\caption{Average PSNR over input band number with scaling factor $\times$3.}
\label{fig:avg_psnr}
\end{figure}


\begin{figure*}[ht]
\begin{center}
\includegraphics[width=\linewidth]{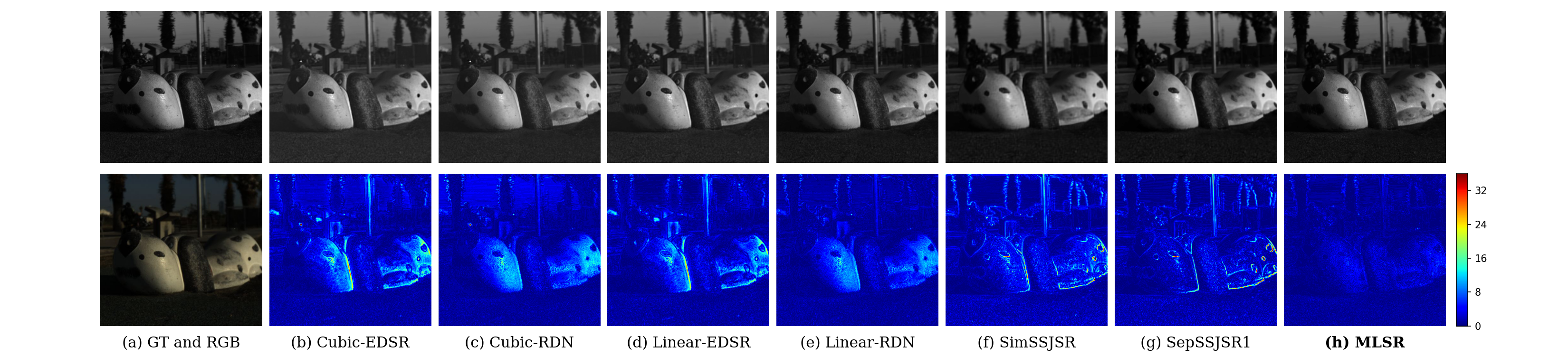}
\includegraphics[width=\linewidth]{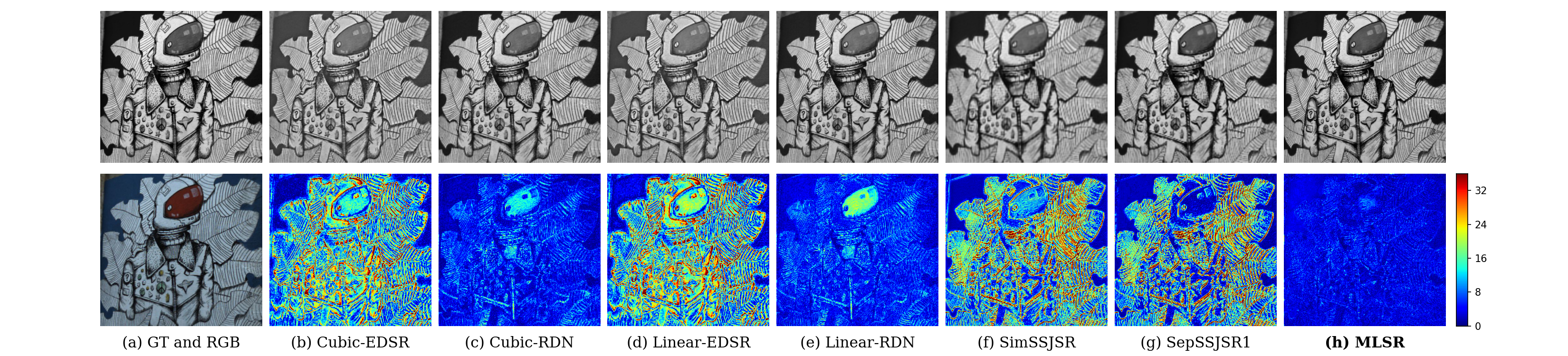}
\end{center}
\vspace{-0.4cm}
\caption{Predicted HSI bands by different methods and their absolute error map. Band 3 is selected in the upper image, band 20 selected in the lower image. The the scaling factor is $\times$3.}
\label{fig:visualization}
\end{figure*}

\begin{figure}[t]
\centering
\begin{minipage}[t]{\columnwidth}
\centering
\setlength{\tabcolsep}{0pt}
\renewcommand{\arraystretch}{0}
\begin{tabular}[c]{c}
    \begin{subfigure}[c]{\columnwidth}
      \includegraphics[width=\columnwidth]{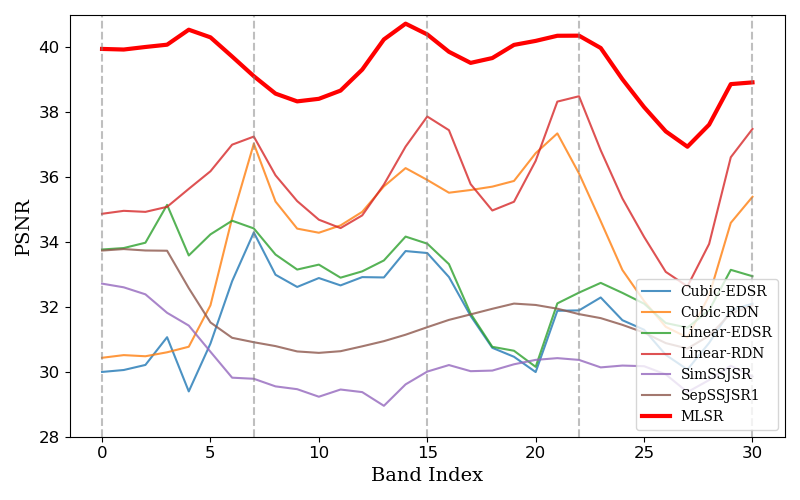}
      \vspace{-0.4cm}
      \caption{The PSNR of different methods across different bands with LR HSIs with 5 input bands. The gray vertical dash lines highlights the band's indices that is part of the input LR HSI.}
      \label{fig:psnr_comb_a}
    \end{subfigure}\\
    
    \begin{subfigure}[c]{\columnwidth}
      \includegraphics[width=\columnwidth]{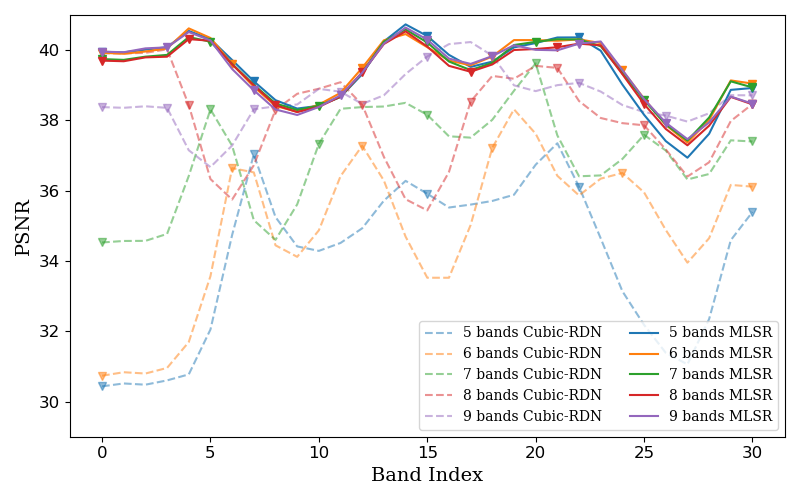}
      \vspace{-0.4cm}
      \caption{Illustration of PSNR curves for different number of input bands for MLSR and Cubic-RDN baseline. The input HSI bands' indices are highlighted by the small triangles with the same color.}
      \label{fig:psnr_comb_b}
    \end{subfigure}
    
    \end{tabular}
  \end{minipage}
  \caption{Performance of MLSR and baselines across different parts of the spectrum with the scaling factor $\times$3.}
\label{fig:psnr_comb}
\vspace{-0.3cm}
\end{figure}

\subsection{Model Performance}
Following the common practice, we evaluate the MLSR on three scales, i.e., $2\times$, $3\times$, $4\times$. For each scale and each dataset, one model is trained and evaluated. For evaluation, peak signal-to-noise ratio (PSNR), and structure similarity (SSIM), which are the two most widely used quantitative picture quality indices (PQIs), are adopted.

Table \ref{tab:results} shows the evaluation results on MLSR and baseline methods. It is clear that for both datasets and under most settings, MLSR gets the same level or better results than baseline models. Fig. \ref{fig:visualization} shows two examples of the resulting SR bands comparison and the corresponding residual error map. The figure clearly shows that though the two baselines proposed by \cite{meiSpatialSpectralJoint2020} are good at predicting low-frequency parts, they are not working well on high-frequency parts like edges, which leads to bad results.

It can be observed that the number of input bands largely influences the performance of baseline methods. With fewer input bands, the spectral-spatial SR becomes more challenging. To evaluate the impact of the input band number on the performance, the input band number varies from 5 to 9 using equidistant sampling. 
As shown in Fig. \ref{fig:avg_psnr}, the average PSNR of the output HR HSIs from MLSR is much higher than the baselines. As the number of input bands decreases, the PSNRs of the baselines decrease rapidly. However, MLSR only drops slightly and can maintain high performance across the numbers of input bands.

\subsection{Auxiliary Experiments: Spectral Extrapolation and Randomized Input Bands}\label{aux}
\begin{table}[ht]
    \centering
    \begin{tabularx}{\columnwidth}{l >{\centering\arraybackslash}X >{\centering\arraybackslash}X}
    \thickhline
                Extrapolate:      & Experiment 1 & Experiment 2 \\
    \hline
        Baseline & 30.26/0.675    & 29.27/0.671    \\
    \hline
        MLSR & \textbf{36.40/0.970}   & \textbf{35.99/0.970}   \\
    \thickhline
    \end{tabularx}
    \caption{The performance (PSNR/SSIM) of baseline and MLSR for spectral extrapolation and spatial SR task in two experimental settings. The scaling factor is $\times$2.} 
    \label{tab:auxiliary}
\vspace{-0.3cm}
\end{table}

Since the proposed task requires the model to understand the spectrum and reconstruct HR bands at requested peak wavelengths, it is necessary to know how the model predicts bands whose peak wavelengths are outside of all the training bands' wavelengths. This part is similar to extrapolation, which is generally considered a more challenging problem than interpolation due to the lack of guidance on both sides \cite{talvitieDistanceBasedInterpolationExtrapolation2015}. Two experiments are designed to evaluate the performance of MLSR on spatial SR with spectral extrapolation on LR HSI. In the first experiment, MLSR is asked to generate the last ten bands using only the first 21 bands of the LR input. The second experiment uses the middle 21 bands of the LR input to generate the five bands on each side of the HR output. Both experiments are done on NTIRE with the scaling factor $\times$2. MLSR is compared with the best performing baseline: the Cubic+RDN, and Table \ref{tab:auxiliary} shows the results. Although compared to spectral interpolation, the performance decreases, MLSR is still much better than the baseline on extrapolation.

\begin{table}[ht]
    \centering
    \begin{tabularx}{\columnwidth}{l >{\centering\arraybackslash}X >{\centering\arraybackslash}X}
    \thickhline
        Input Bands Index & Baseline & MLSR \\
    \hline
        2,5,11,20,28 & 29.75/0.860            & \tb{38.64/0.976} \\
    \hline
        7,19,21,25,29 & 23.57/0.649           & \tb{38.38/0.975} \\
    \hline
        8,12,15,20,21 & 22.27/0.437           & \tb{38.02/0.973} \\
    \hline
        1,4,9,12,19,28 & 30.91/0.875          & \tb{38.63/0.976} \\
    \hline
        5,6,8,17,19,20,25 & 28.09/0.663       & \tb{38.36/0.975} \\
    \hline
        6,7,9,10,11,12,15,17,22 & 24.53/0.587 & \tb{38.07/0.973} \\
    
    \thickhline
    \end{tabularx}
    \caption{The performance (PSNR/SSIM) of baseline and MLSR with randomly selected HSI bands. The scaling factor is $\times2$, and the baseline method is Cubic+RDN.}
    \label{tab:random}
\vspace{-0.3cm}
\end{table}

    
Another experiment is conducted to evaluate further the model's ability to perform spectral HR reconstruction on more diverse input configurations. The model outputs the whole 31 bands with randomly selected 5 to 9 bands from the spectrum of LR HSI as input. With the random selection, the input bands can cluster together and can be unevenly distributed. As a result, interpolation and extrapolation are conducted at the same time. As shown in Table \ref{tab:random}, the performance of MLSR is consistently better than the baseline on all randomly selected input formats with the scaling factor $\times$2. Moreover, a single MLSR model to perform all the experiments in Table \ref{tab:random} is trained. The results prove that the proposed W2WNet can learn the knowledge embedded in the spectrum instead of the statistical bias in the predefined wavelengths and settings.

\subsection{Performance over Output Bands}

So far, it is well proved that the total number of input bands impacts the performance. In this subsection, how the input bands' distribution affect the output bands is studied.
Fig. \ref{fig:psnr_comb_a} shows the band-wise PSNR of MLSR and baselines with five input bands. The scaling factor is set to $\times$3. The baseline models find local maxima at the same bands used in the input (highlighted by the vertical dash line in Fig. \ref{fig:psnr_comb_a}). As the distance to the nearest input band index increases, there is a fairly sharp decrease in the baseline models' performance. Overall, the performances of the baseline models are dictated by the input LR HSI bands' locations. The proposed MLSR consistently outperforms all the baseline models across all the bands. While the difference in performance between MLSR and the baseline models is relatively low at the input bands' indices, the main advantage of MLSR can be observed at indices further away from these bands. Overall the proposed MLSR gives a relatively more flat response curve and can generate superior spectral-spatial super-resolved HSI at intermediate wavelengths.

Fig. \ref{fig:psnr_comb_b} shows the PSNR of MLSR and one top-performing baseline (i.e., Cubic RDN) across all the bands as the number of input band number changes with the scaling factor $\times$3. Again, the performance of the baseline model is primarily dictated by choice of the input bands, and the shape of the PSNR curve varies significantly as the input band number changes. Surprisingly, the shape of the PSNR curve of the proposed MLSR model almost remains the same as the number of input bands and the input band indices changes. Overall, it indicates that MLSR is agnostic to the input band settings and provides a consistent, flatter, and higher response across the spectrum.


\subsection{Ablation Study}

To prove the effectiveness of W2WNet and Meta-WEC, Meta-WECs are replaced with an ordinary $1\times 1$ convolution, removes the W2WNet, and keeps all other parts the same. The scaling factor is set to $\times$2, the input bands number is five, and the dataset is NTIRE2020. The first experiment in Table \ref{tab:ablation} only replaces the Meta-WEC in SOFE, while the second replaces Meta-WECs in both SOFE and SOBR.
Also, to prove that $1\times1$ kernel is a better choice for Meta-WEC than other settings like widely-used $3\times3$, the same experiment as above is done with the Meta-WEC kernel's size set to $3\times3$.
\begin{table}[ht]
    \centering
    \begin{tabularx}{\columnwidth}{l >{\centering\arraybackslash}X >{\centering\arraybackslash}X}
    \thickhline
        Experiment & Results \\
    \hline
        Replace Meta-WEC in SOFE & 39.13/0.978  \\
    \hline
        Replace Meta-WEC in SOFE and SOBR & 27.76/0.917 \\
    \hline
        Replace Meta-WEC 1$\times$1 kernel by 3$\times$3 & 39.93/0.980 \\
    \hline
        MLSR Original & \tb{41.77/0.985} \\
    \thickhline
    \end{tabularx}
    \caption{Ablation study results(PSNR/SSIM).}
    \label{tab:ablation}
\vspace{-0.3cm}
\end{table}


\section{Conclusion}


This paper proposes a Meta fusion-based framework for spectral understanding, retrieval, and high-resolution HSI reconstruction. The proposed W2WNet and Meta-WEC modules adaptively learn spectral information in each image band and enrich its feature map. Our model has shown significant flexibility and robustness in generating arbitrary input-output bands within the same domain knowledge datasets. It is the first step toward cross-dataset HSI SR with diverse domain knowledge. Extensive experiments show the superiority of our novel design.

{\small
\bibliographystyle{ieee_fullname}
\bibliography{reference}
}

\end{document}